\documentclass[acmtog,review=false, nonacm]{acmart}
\usepackage{booktabs} 
\usepackage{import}

\def\fg#1{Fig.~\ref{fig:#1}}
\def\sec#1{Section~\ref{sec:#1}}

\graphicspath{ {./_figs/} }

\begin{document}

\title{Real Image Inversion via Segments}

\author{David Futschik}
\email{futscdav [at] fel.cvut.cz}

\affiliation{%
  \institution{Czech Technical University in Prague, Faculty of Electrical Engineering}
  \streetaddress{Karlovo n\'{a}m\v{e}st\'{i} 13}
  \city{Praha 2}
  \state{Czech Republic}
  \postcode{121 35}
}

\author{Michal Luk\'{a}\v{c}}
\email{elishe [at] adobe.com}

\affiliation{%
  \institution{Adobe Research}
  \streetaddress{345 Park Ave}
  \city{San Jose}
  \state{CA}
  \postcode{95110}
}

\author{Eli Shechtman}
\email{elishe [at] adobe.com}

\affiliation{%
  \institution{Adobe Research}
  \streetaddress{801 N 34th St}
  \city{Seattle}
  \state{WA}
  \postcode{98103}
}

\author{Daniel S\'{y}kora}
\email{sykorad (at) fel.cvut.cz}

\affiliation{%
  \institution{Czech Technical University in Prague, Faculty of Electrical Engineering}
  \streetaddress{Karlovo n\'{a}m\v{e}st\'{i} 13}
  \city{Praha 2}
  \state{Czech Republic}
  \postcode{121 35}
}

\begin{teaserfigure}
\def\svgwidth{\linewidth}
\begingroup%
  \makeatletter%
  \providecommand\color[2][]{%
    \errmessage{(Inkscape) Color is used for the text in Inkscape, but the package 'color.sty' is not loaded}%
    \renewcommand\color[2][]{}%
  }%
  \providecommand\transparent[1]{%
    \errmessage{(Inkscape) Transparency is used (non-zero) for the text in Inkscape, but the package 'transparent.sty' is not loaded}%
    \renewcommand\transparent[1]{}%
  }%
  \providecommand\rotatebox[2]{#2}%
  \ifx\svgwidth\undefined%
    \setlength{\unitlength}{1662.4bp}%
    \ifx\svgscale\undefined%
      \relax%
    \else%
      \setlength{\unitlength}{\unitlength * \real{\svgscale}}%
    \fi%
  \else%
    \setlength{\unitlength}{\svgwidth}%
  \fi%
  \global\let\svgwidth\undefined%
  \global\let\svgscale\undefined%
  \makeatother%
  \begin{picture}(1,0.52697454)%
    \put(0,0){\includegraphics[width=\unitlength]{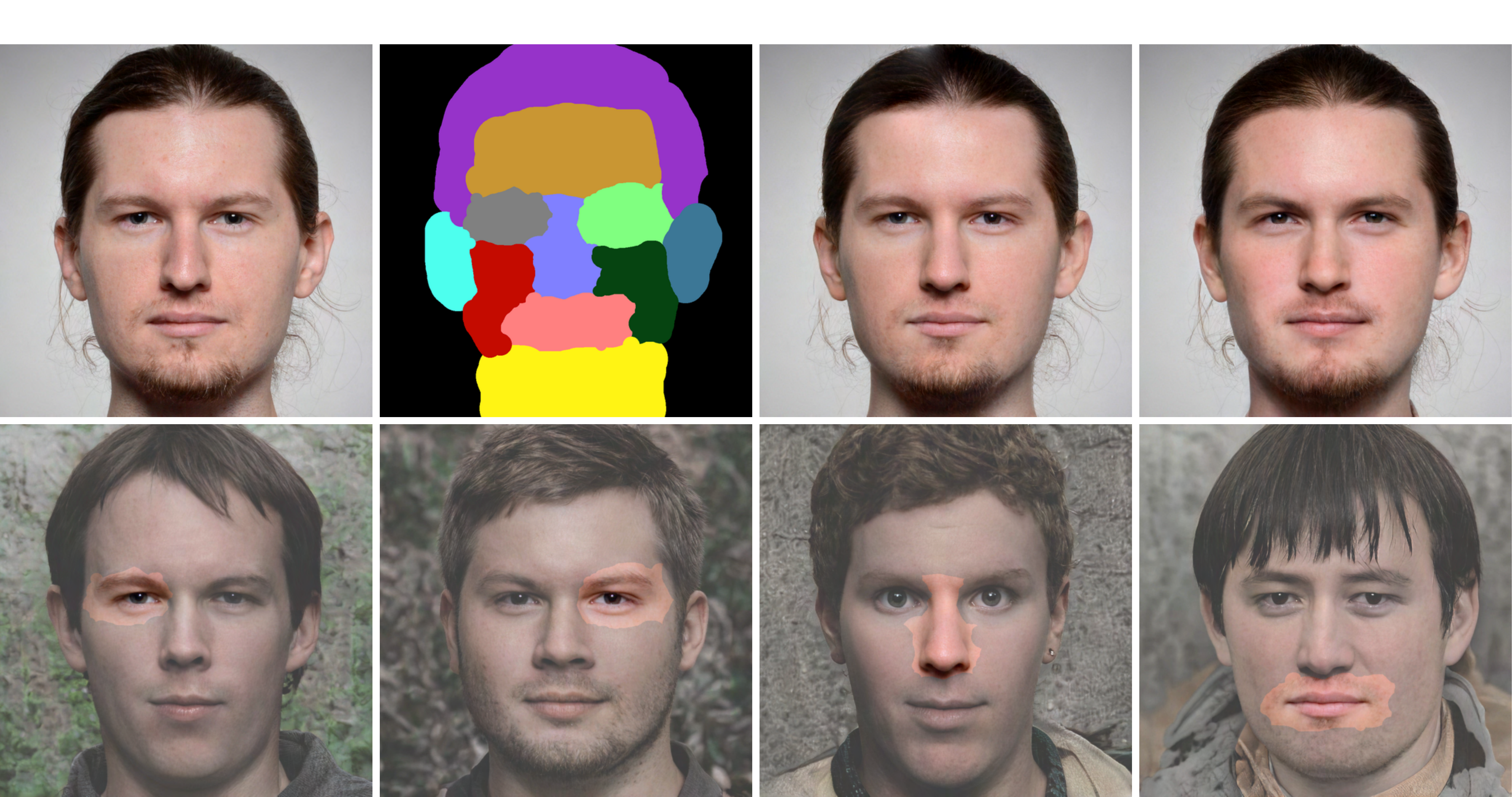}}%
    \put(0.12320665,0.510){\color[rgb]{0,0,0}\makebox(0,0)[b]{\smash{(a)~Original image}}}%
    \put(0.37440973,0.510){\color[rgb]{0,0,0}\makebox(0,0)[b]{\smash{(b)~Segments}}}%
    \put(0.62561281,0.510){\color[rgb]{0,0,0}\makebox(0,0)[b]{\smash{(c)~Segmented projection}}}%
    \put(0.87681589,0.510){\color[rgb]{0,0,0}\makebox(0,0)[b]{\smash{(d)~Global projection}}}%
    \put(0.01084738,0.225){\color[rgb]{0,0,0}\makebox(0,0)[lb]{\smash{(e)}}}%
    \put(0.26172963,0.225){\color[rgb]{0,0,0}\makebox(0,0)[lb]{\smash{(f)}}}%
    \put(0.51261190,0.225){\color[rgb]{0,0,0}\makebox(0,0)[lb]{\smash{(g)}}}%
    \put(0.76349416,0.225){\color[rgb]{0,0,0}\makebox(0,0)[lb]{\smash{(h)}}}%
  \end{picture}%
\endgroup%

\caption{A key concept of our method: The original photo~(a) is subdivided
into a set of segments~(b) for each of which projection into a latent space
is peformed indenepndently~(e--h), i.e., each segment has its own latent code.
Thanks to the lower number of constraints used for projection in each segment
the resulting stitched image~(c) preserves the identity notably better when
compared to a global projection with a single latent code~(d).}
\label{fig:segments}
\end{teaserfigure}
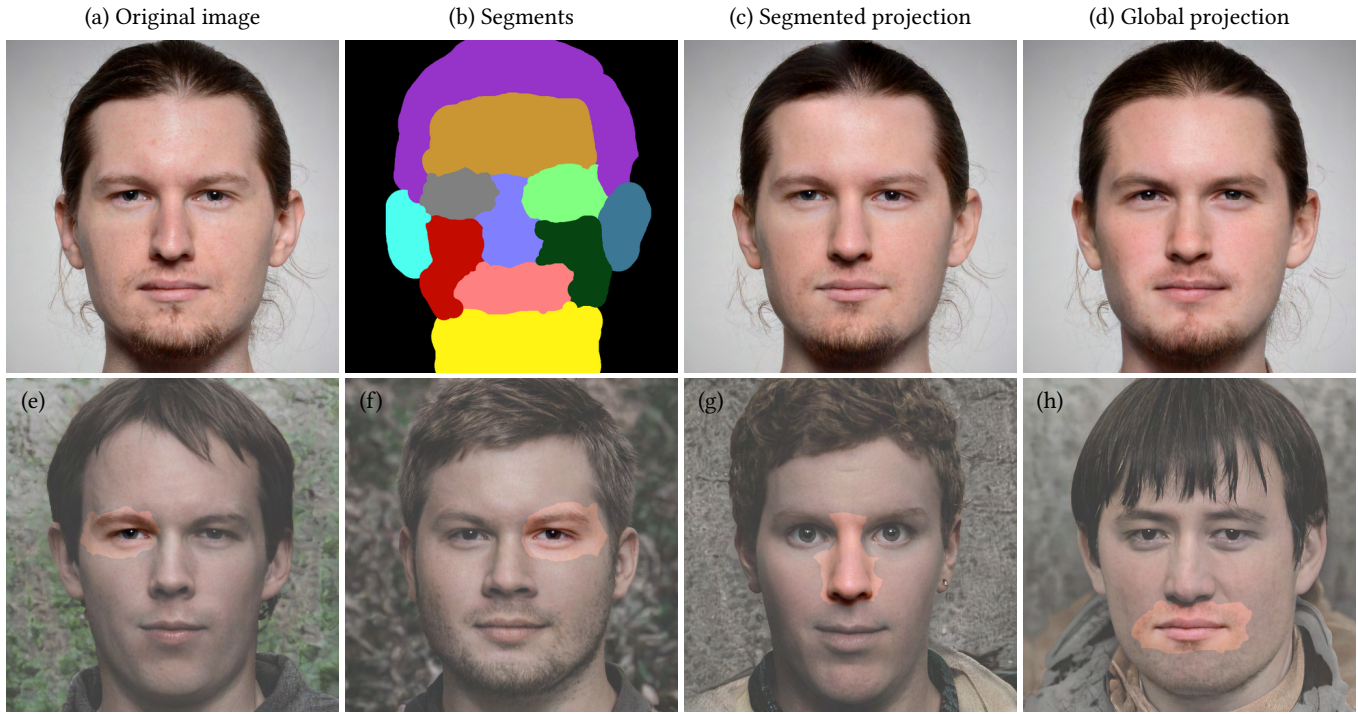

\begin{abstract}

In this short report, we present a simple, yet effective approach to editing
real images via generative adversarial networks (GAN). Unlike previous
techniques, that treat all editing tasks as an operation that affects pixel
values in the entire image in our approach we cut up the image into a set of
smaller segments. For those segments corresponding latent codes of a generative
network can be estimated with greater accuracy due to the lower number of
constraints. When codes are altered by the user the content in the image is
manipulated locally while the rest of it remains unaffected. Thanks to this
property the final edited image better retains the original structures and thus
helps to preserve natural look.

\end{abstract}

\maketitle

\section{Introduction}

With the increasing ability of GANs to generate images hardly distinguishable
from real photographs, there have been numerous attempts to estimate latent
code of the network generating images that look very close to the given input
photo. By manipulating those codes in a specific direction one may alter the
appearance of the input photo in a specific way while retaining the original
visual features, e.g., adding more hair to a bald person while retaining their
identity.

A large body of work has been done to find an intuitive projection of latent
codes into to a space in which important visual features will be disentangled
so that the user can perform edits locally. Unfortunately, there is usually a
trade off between the ability to get accurate reconstruction while at the same
time provide an excellent disentanglement. For example in StyleGAN
v2~\cite{Karras20}, the original input code~$z\in \mathbb{R}^{512}$ is
transformed into a vector~$\mathcal{W}\in \mathbb{R}^{512}$ which is easy to
edit, however, difficult to estimate. While in a different approach of Karras
et al.~\shortcite{Karras19}~$\mathcal{W}^{+}\in \mathbb{R}^{18\times512}$ is
used that has enough degrees of freedom to provide a good code estimation,
nevertheless, it is more difficult to manipulate.

A key obstacle which stays behind this trade off is the fact that previous
techniques work with the assumption that the entire image is represented by a
single latent code. However, considering all pixels in the input image imposes
a large number of constraints, which can make the code estimation difficult
(see~\fg{segments}d). In our solution we relax this assumption and instead of
using one single latent code, we segment the input image into a set of regions
(\fg{segments}b) for which we estimate local codes separately
(\fg{segments}e--h). Then, to reconstruct the image, we generate each segment
from the corresponding latent code separately and compose the final image using
those pieces~(\fg{segments}c).

Besides the lower number of constraints that enables estimation of latent codes
that generate images closer to the input photo our segmentation-based technique
also ensures precise localization of subsequent edits. Thanks to that property
important visual features remain intact and thus help to retain fidelity of the
original image.

The contributions of this short report are the following: (1)~We demonstrate
how to perform segmentation-based estimation of latent codes and how this
approach can improve the ability to reconstruct important visual features in
the input image. (2)~We show various use cases in which precise localisation
enable alternations that would be difficult to achieve using current
state-of-the-art.

\section{Method}

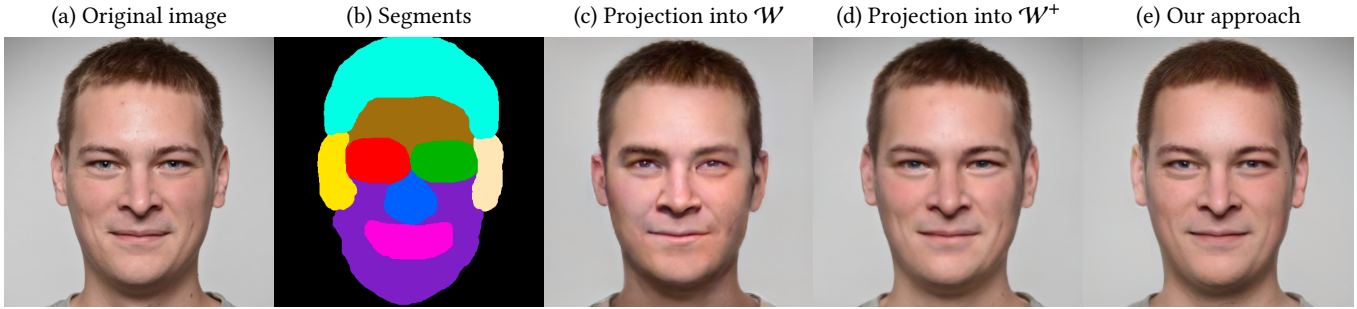
\begin{figure*}[ht]
\def\svgwidth{\linewidth}
\begingroup%
  \makeatletter%
  \providecommand\color[2][]{%
    \errmessage{(Inkscape) Color is used for the text in Inkscape, but the package 'color.sty' is not loaded}%
    \renewcommand\color[2][]{}%
  }%
  \providecommand\transparent[1]{%
    \errmessage{(Inkscape) Transparency is used (non-zero) for the text in Inkscape, but the package 'transparent.sty' is not loaded}%
    \renewcommand\transparent[1]{}%
  }%
  \providecommand\rotatebox[2]{#2}%
  \newcommand*\fsize{\dimexpr\f@size pt\relax}%
  \newcommand*\lineheight[1]{\fontsize{\fsize}{#1\fsize}\selectfont}%
  \ifx\svgwidth\undefined%
    \setlength{\unitlength}{1919.99994233bp}%
    \ifx\svgscale\undefined%
      \relax%
    \else%
      \setlength{\unitlength}{\unitlength * \real{\svgscale}}%
    \fi%
  \else%
    \setlength{\unitlength}{\svgwidth}%
  \fi%
  \global\let\svgwidth\undefined%
  \global\let\svgscale\undefined%
  \makeatother%
  \begin{picture}(1,0.22348188)%
    \lineheight{1}%
    \setlength\tabcolsep{0pt}%
    \put(0,0){\includegraphics[width=\unitlength,page=1]{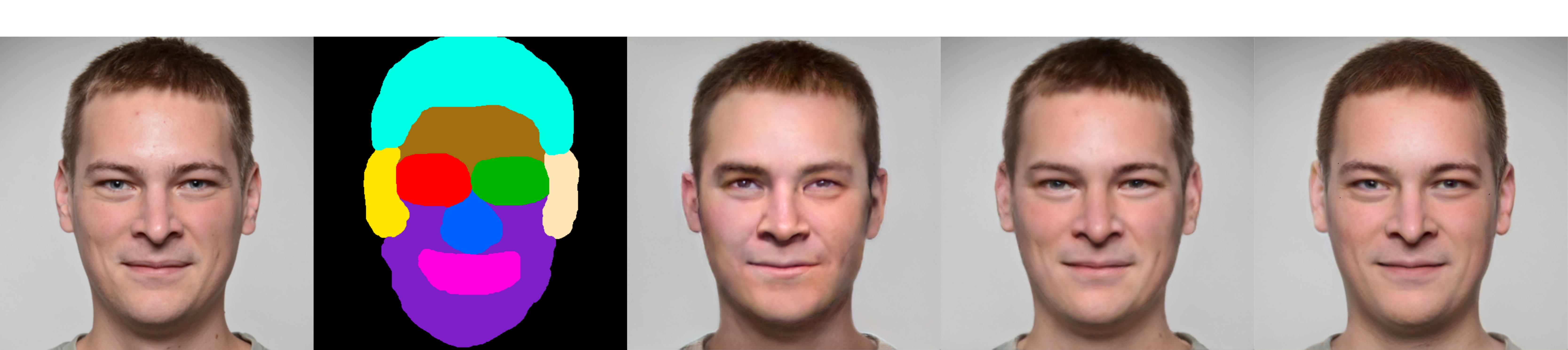}}%
    \put(0.1,0.211){\color[rgb]{0,0,0}\makebox(0,0)[t]{\lineheight{1.25}\smash{\begin{tabular}[t]{c}(a)~Original image\end{tabular}}}}%
    \put(0.3,0.211){\color[rgb]{0,0,0}\makebox(0,0)[t]{\lineheight{1.25}\smash{\begin{tabular}[t]{c}(b)~Segments\end{tabular}}}}%
    \put(0.5,0.211){\color[rgb]{0,0,0}\makebox(0,0)[t]{\lineheight{1.25}\smash{\begin{tabular}[t]{c}(c)~Projection into~$\mathcal{W}$\end{tabular}}}}%
    \put(0.7,0.211){\color[rgb]{0,0,0}\makebox(0,0)[t]{\lineheight{1.25}\smash{\begin{tabular}[t]{c}(d)~Projection into~$\mathcal{W}^{+}$\end{tabular}}}}%
    \put(0.9,0.211){\color[rgb]{0,0,0}\makebox(0,0)[t]{\lineheight{1.25}\smash{\begin{tabular}[t]{c}(e)~Our approach\end{tabular}}}}%
    \put(-0.03077809,0.30778087){\color[rgb]{0,0,0}\makebox(0,0)[lt]{\begin{minipage}{1.04645489\unitlength}\raggedright \end{minipage}}}%
  \end{picture}%
\endgroup%

\caption{Comparison of identity preservation after projecting the original
photo~(a) into~$\mathcal{W}$~(c), $\mathcal{W}^{+}$~(d), and using our
approach~(e) with segments~(b).}\label{fig:inversion}
\end{figure*}

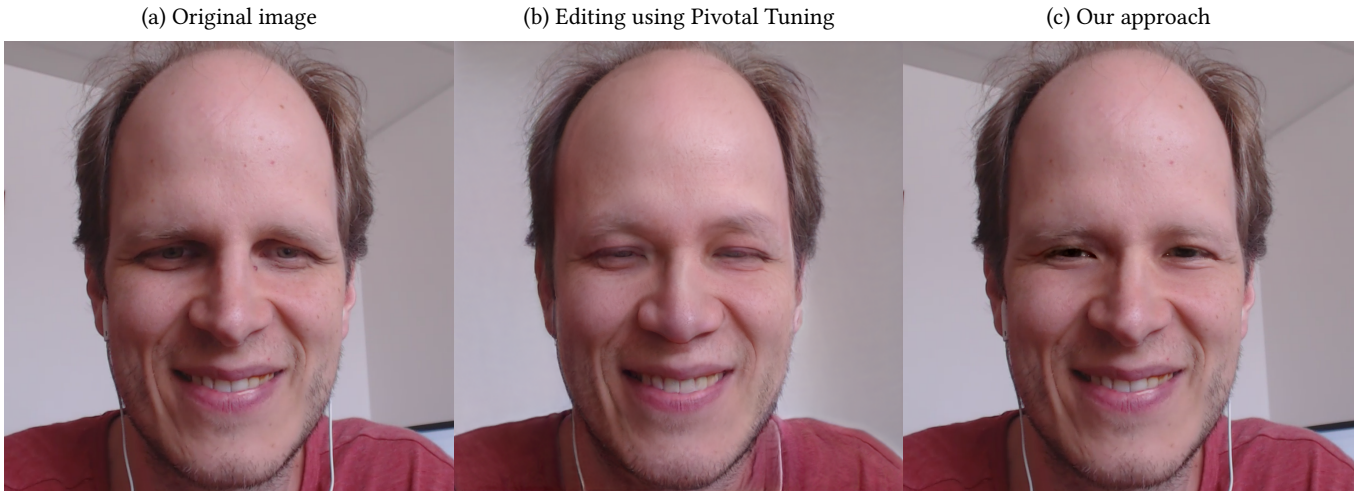
\begin{figure*}
\def\svgwidth{\linewidth}
\begingroup%
  \makeatletter%
  \providecommand\color[2][]{%
    \errmessage{(Inkscape) Color is used for the text in Inkscape, but the package 'color.sty' is not loaded}%
    \renewcommand\color[2][]{}%
  }%
  \providecommand\transparent[1]{%
    \errmessage{(Inkscape) Transparency is used (non-zero) for the text in Inkscape, but the package 'transparent.sty' is not loaded}%
    \renewcommand\transparent[1]{}%
  }%
  \providecommand\rotatebox[2]{#2}%
  \newcommand*\fsize{\dimexpr\f@size pt\relax}%
  \newcommand*\lineheight[1]{\fontsize{\fsize}{#1\fsize}\selectfont}%
  \ifx\svgwidth\undefined%
    \setlength{\unitlength}{2303.99979238bp}%
    \ifx\svgscale\undefined%
      \relax%
    \else%
      \setlength{\unitlength}{\unitlength * \real{\svgscale}}%
    \fi%
  \else%
    \setlength{\unitlength}{\svgwidth}%
  \fi%
  \global\let\svgwidth\undefined%
  \global\let\svgscale\undefined%
  \makeatother%
  \begin{picture}(1,0.36258795)%
    \lineheight{1}%
    \setlength\tabcolsep{0pt}%
    \put(0,0){\includegraphics[width=\unitlength,page=1]{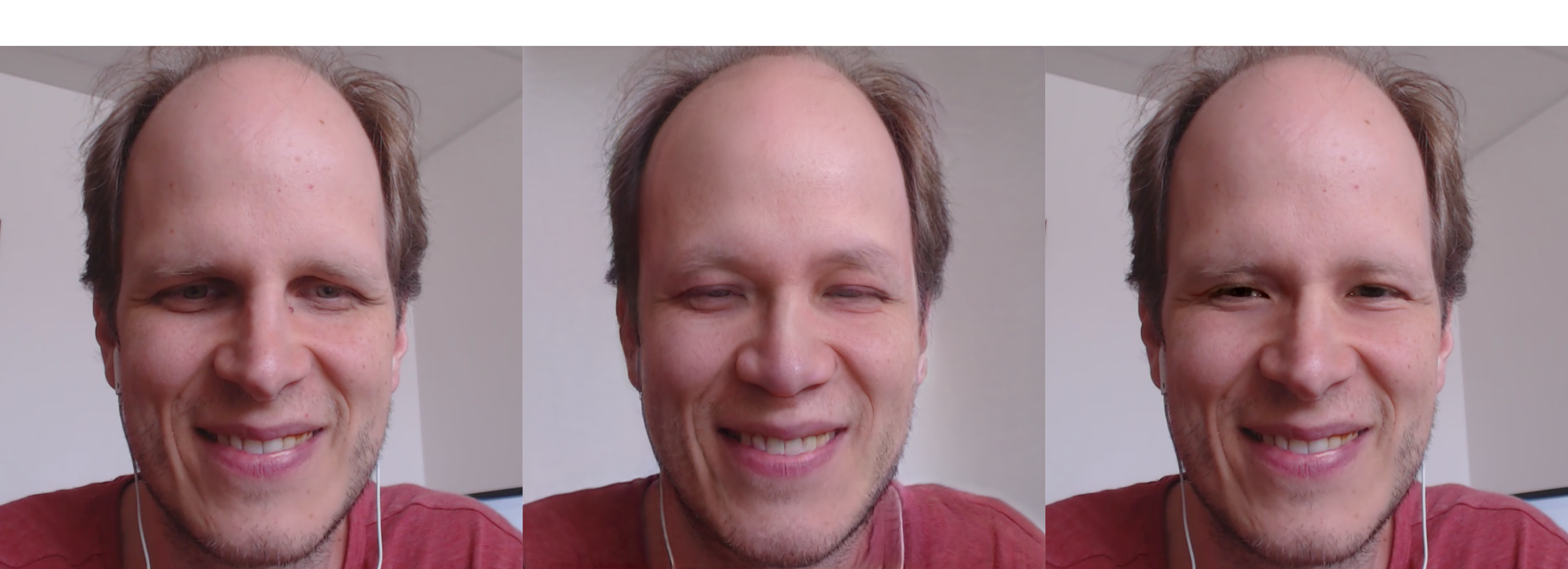}}%
    \put(0.16666984,0.34586154){\color[rgb]{0,0,0}\makebox(0,0)[t]{\lineheight{1.25}\smash{\begin{tabular}[t]{c}(a)~Original image\end{tabular}}}}%
    \put(0.5000032,0.34586154){\color[rgb]{0,0,0}\makebox(0,0)[t]{\lineheight{1.25}\smash{\begin{tabular}[t]{c}(b)~Editing using Pivotal Tuning\end{tabular}}}}%
    \put(0.83333653,0.34586154){\color[rgb]{0,0,0}\makebox(0,0)[t]{\lineheight{1.25}\smash{\begin{tabular}[t]{c}(c)~Our approach\end{tabular}}}}%
  \end{picture}%
\endgroup%

\caption{Comparison of editing in $\mathcal{S}$-space~\cite{Wu21} using Pivotal
Tuning~\cite{Roich21}~(b) and our approach~(c). Note, how even with the Pivotal
Tuning and disentangled feature editing space it is still difficult to achieve
good identity preservation of the original image~(a).}\label{fig:pivotal_tuning}
\end{figure*}

\subsection{Projection}

Given a real image~$I$ and a set of segments~$S_{1, \ldots, n}$ which
correspond to regions in~$I$, we perform~$n$ separate projections of~$S_{n}(I)$
into the desired input latent space~$X_n$. The domain of~$X$ is not necessarily
fixed. In general it can be any input space that can compactly represent inputs
into the GAN. In the case of StyleGAN v2~\cite{Karras20}, we consider~$z$,
$\mathcal{W}$, $\mathcal{W}^{+}$, and $\mathcal{S}$-space~\cite{Wu21}, however,
any previously published or a newly developed projection method can be used. In
fact our segmented projection is a complementary improving mechanism that can
help to achieve better results regardless the selected projection method
(see~\fg{inversion}). Those can possibly be even mixed to achieve best
results.

The function of~$S(I)$ depends on the method which provides the input based
on~$I$. In the case of gradient descent on some image loss~$\mathcal{L}$,
$S(I)$~can be a composition of functions which combine the region of interest
defined by~$S_k$ with~$I$ such that per-pixel loss outside the region is equal
to~$0$. If we want to use a feed-forward code estimation, we provide the mask
as a part of the input and alter the training regime of the code-providing
network.

To minimize seams between individual projected segments we dilate their
boundaries to achieve spatial overlap. Alternatively we can condition the projection
loss so that it considers also a small neighborhood around each segment. When
there are areas that we do not wish to edit at all, we exclude them from the
projection altogether and in the final composition we will substitute them back
directly from~$I$.

To achieve even better inversion, we can further fine tune the underlying model
for each segment separately using state-of-the-art techniques such as Pivotal
Tuning~\cite{Roich21}~(see~\fg{pivotal_tuning}). Like in the projection task,
fine tuning is made considerably easier with reduced number of constraints
offered by only considering the region of interest.

\begin{figure*}[ht]
\def\svgwidth{\linewidth}
\begingroup%
  \makeatletter%
  \providecommand\color[2][]{%
    \errmessage{(Inkscape) Color is used for the text in Inkscape, but the package 'color.sty' is not loaded}%
    \renewcommand\color[2][]{}%
  }%
  \providecommand\transparent[1]{%
    \errmessage{(Inkscape) Transparency is used (non-zero) for the text in Inkscape, but the package 'transparent.sty' is not loaded}%
    \renewcommand\transparent[1]{}%
  }%
  \providecommand\rotatebox[2]{#2}%
  \newcommand*\fsize{\dimexpr\f@size pt\relax}%
  \newcommand*\lineheight[1]{\fontsize{\fsize}{#1\fsize}\selectfont}%
  \ifx\svgwidth\undefined%
    \setlength{\unitlength}{14916.81625246bp}%
    \ifx\svgscale\undefined%
      \relax%
    \else%
      \setlength{\unitlength}{\unitlength * \real{\svgscale}}%
    \fi%
  \else%
    \setlength{\unitlength}{\svgwidth}%
  \fi%
  \global\let\svgwidth\undefined%
  \global\let\svgscale\undefined%
  \makeatother%
  \begin{picture}(1,0.25008455)%
    \lineheight{1}%
    \setlength\tabcolsep{0pt}%
    \put(0,0){\includegraphics[width=\unitlength,page=1]{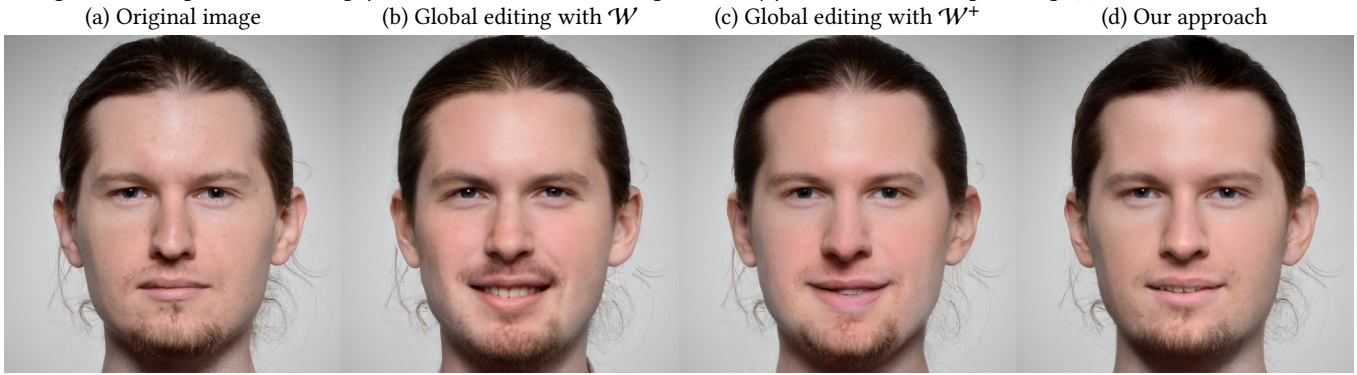}}%
    \put(0.12502721,0.25869609){\color[rgb]{0,0,0}\makebox(0,0)[t]{\lineheight{1.25}\smash{\begin{tabular}[t]{c}(a)~Original image\end{tabular}}}}%
    \put(0.37490367,0.25869614){\color[rgb]{0,0,0}\makebox(0,0)[t]{\lineheight{1.25}\smash{\begin{tabular}[t]{c}(b)~Global editing with~$\mathcal{W}$\end{tabular}}}}%
    \put(0.62499805,0.25869082){\color[rgb]{0,0,0}\makebox(0,0)[t]{\lineheight{1.25}\smash{\begin{tabular}[t]{c}(c)~Global editing with~$\mathcal{W}^{+}$\end{tabular}}}}%
    \put(0.87480818,0.25869609){\color[rgb]{0,0,0}\makebox(0,0)[t]{\lineheight{1.25}\smash{\begin{tabular}[t]{c}(d)~Our approach\end{tabular}}}}%
  \end{picture}%
\endgroup%

\caption{Comparison of editing in $\mathcal{S}$-space~\cite{Wu21}: While
changes of~$\mathcal{W}$ in $\mathcal{S}$-space can produce naturally looking
results~(b), the identity is quite far from the original image~(a). Identity
can be preserved notably better with~$\mathcal{W}^{+}$, however, the result
after performing a similar edit is not believable~(c). Using our method~(d) one
can obtain better identity preservation as well as naturally looking edits (all
segments are changed simultaneously).}
\label{fig:smile}
\end{figure*}

\begin{figure*}
\def\svgwidth{\linewidth}\input{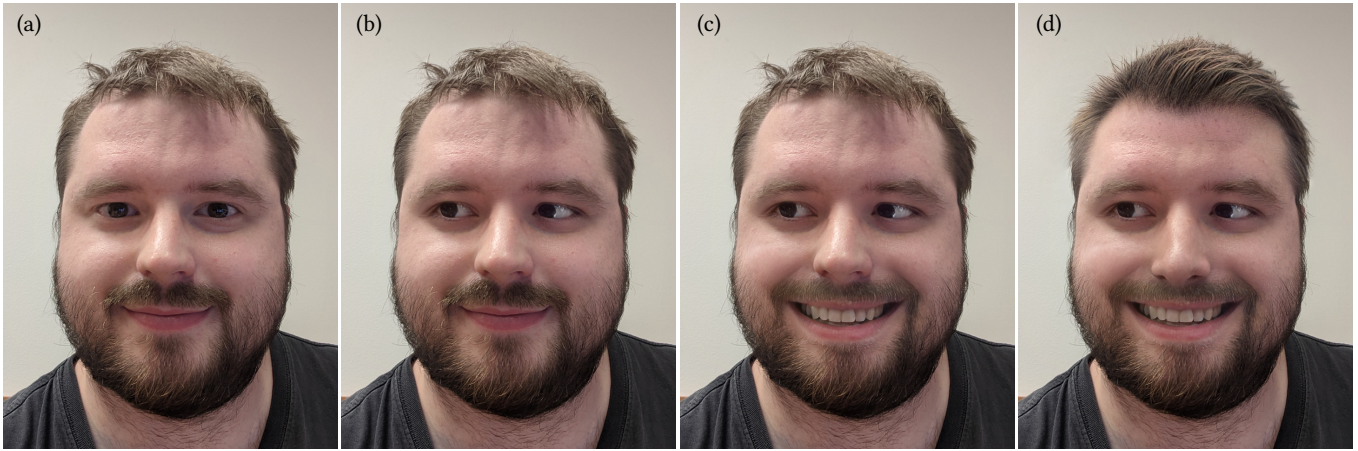}
\caption{Examples of local layered edits applied subsequently on a real
photograph~(a): changing gaze direction~(b), adding smile~(c), changing haircut
and nose shape~(d).}\label{fig:edits}
\end{figure*}

\subsection{Editing}

Once the projection is ready, we can perform latent space edits to achieve
natural-looking changes in the input image. To achieve best results, we
prepare~$S$ to follow semantic segmentation, i.e., segments containing
individual facial feature (see, e.g., masks show in~\fg{segments}
and~\fg{inversion}).

First type of edits we can perform are locally global, i.e., we change all
segments in the same direction. For example, we can perform
various~$\mathcal{W}^k_{edited} = \mathcal{W}^k + \alpha D$ changes, with known
directions~$D$, and then compose the final image using~$S$. This type of edit
can be seen in~\fg{smile}.

Another possibility is to perform incremental modification. In this scenario
segments are optimized sequentially, edited, and the final composite becomes
again a new~$I$ for the next iteration. This editing strategy is shown
in~\fg{edits}. Such a workflow is intuitive for users as they can exactly
specify what they want to change, overview the resulting composition, and then
possibly go back and revise their requirements by making changes in other
segments.

When making a final composition of an edited image, even when the edits are
consistent in all segments global continuity is no longer guaranteed and thus
a small correction is usually necessary. To this end, we employ Poisson image
stitching~\cite{Perez03} between the segments to hide minor discrepancies.

\begin{figure*}
\def\svgwidth{\linewidth}
\begingroup%
  \makeatletter%
  \providecommand\color[2][]{%
    \errmessage{(Inkscape) Color is used for the text in Inkscape, but the package 'color.sty' is not loaded}%
    \renewcommand\color[2][]{}%
  }%
  \providecommand\transparent[1]{%
    \errmessage{(Inkscape) Transparency is used (non-zero) for the text in Inkscape, but the package 'transparent.sty' is not loaded}%
    \renewcommand\transparent[1]{}%
  }%
  \providecommand\rotatebox[2]{#2}%
  \newcommand*\fsize{\dimexpr\f@size pt\relax}%
  \newcommand*\lineheight[1]{\fontsize{\fsize}{#1\fsize}\selectfont}%
  \ifx\svgwidth\undefined%
    \setlength{\unitlength}{3071.99983852bp}%
    \ifx\svgscale\undefined%
      \relax%
    \else%
      \setlength{\unitlength}{\unitlength * \real{\svgscale}}%
    \fi%
  \else%
    \setlength{\unitlength}{\svgwidth}%
  \fi%
  \global\let\svgwidth\undefined%
  \global\let\svgscale\undefined%
  \makeatother%
  \begin{picture}(1,0.27364886)%
    \lineheight{1}%
    \setlength\tabcolsep{0pt}%
    \put(0,0){\includegraphics[width=\unitlength,page=1]{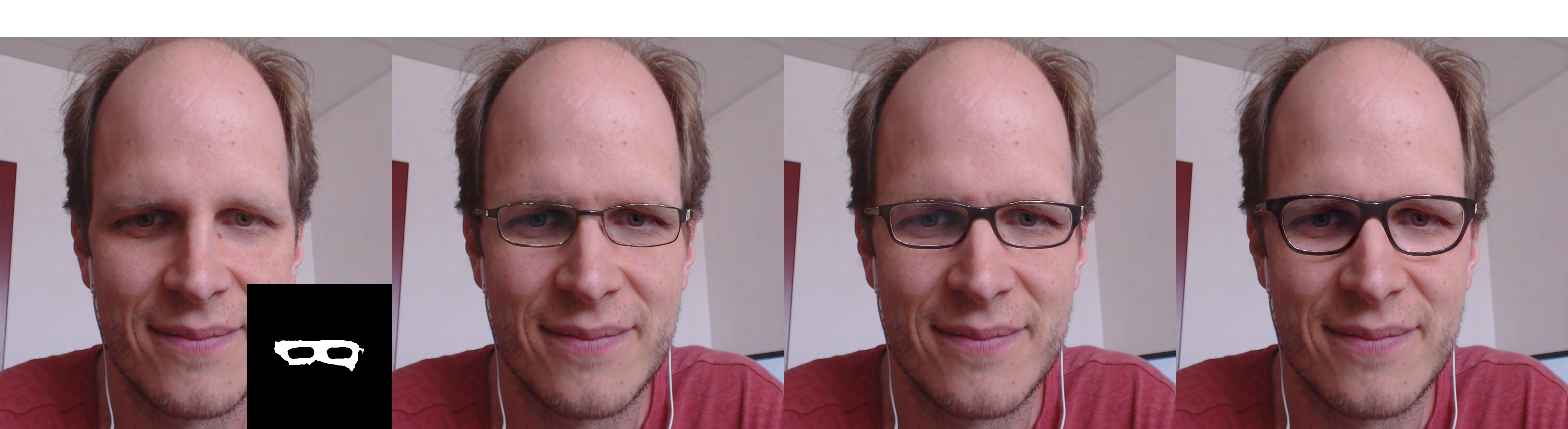}}%
    \put(0.12500239,0.26412421){\color[rgb]{0,0,0}\makebox(0,0)[t]{\lineheight{1.25}\smash{\begin{tabular}[t]{c}(a)\end{tabular}}}}%
    \put(0.37500238,0.26412421){\color[rgb]{0,0,0}\makebox(0,0)[t]{\lineheight{1.25}\smash{\begin{tabular}[t]{c}(c)\end{tabular}}}}%
    \put(0.62500238,0.26412421){\color[rgb]{0,0,0}\makebox(0,0)[t]{\lineheight{1.25}\smash{\begin{tabular}[t]{c}(d)\end{tabular}}}}%
    \put(0.87500238,0.26412421){\color[rgb]{0,0,0}\makebox(0,0)[t]{\lineheight{1.25}\smash{\begin{tabular}[t]{c}(e)\end{tabular}}}}%
    \put(0.16825017,0.07749507){\color[rgb]{1,1,1}\makebox(0,0)[t]{\lineheight{1.25}\smash{\begin{tabular}[t]{c}(b)\end{tabular}}}}%
  \end{picture}%
\endgroup%

\caption{Our approach~(b) can achieve state-of-the-art identity preservation
even for complex edits such as putting on glasses~(c--e) to a subject in the
original photo~(a).}\label{fig:glasses}
\end{figure*}


\begin{figure*}[ht]
\def\svgwidth{\linewidth}\input{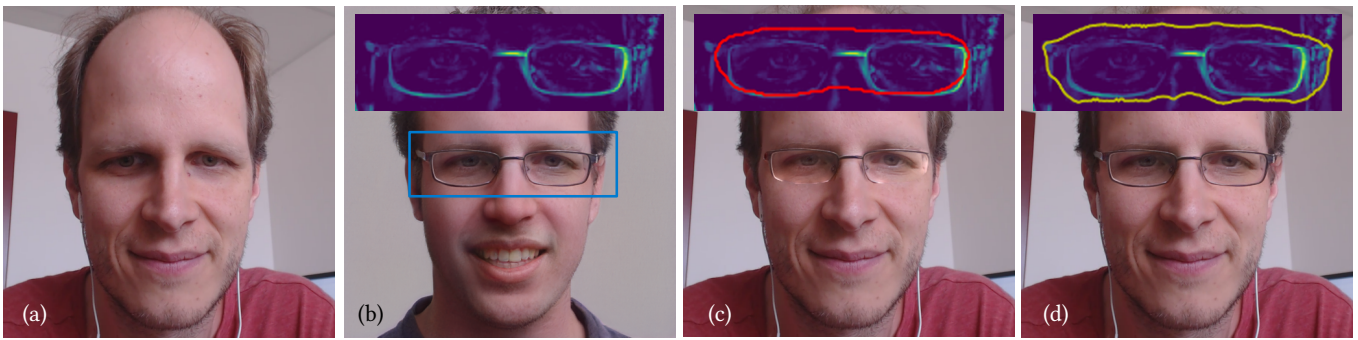}
\caption{An example of automatic mask refinement---our approach is applied to
find and edit a latent code~$\mathcal{W}$ with the aim to add glasses to the
original photo~(a). For the initial projection segmentaiton mask~$S$ denoted by
red curve in the inset of~(c) was used. However, after editing boundary of~$S$
started to collide with new features in the edited image and thus the resulting
composite contains artifacts~(c). Automatic mask refinement (yellow curve) was
used to avoid such collision and produce clean composite~(d). As a stopping
funcion~$F$ for mask refinement based on Level Set method magnitude of pixel
differences between images~(a) and~(b) is used, see insets in figures~(b--d).}
\label{fig:lsm}
\end{figure*}

\begin{figure*}[ht]
\centerline{\def\svgwidth{\linewidth}\input{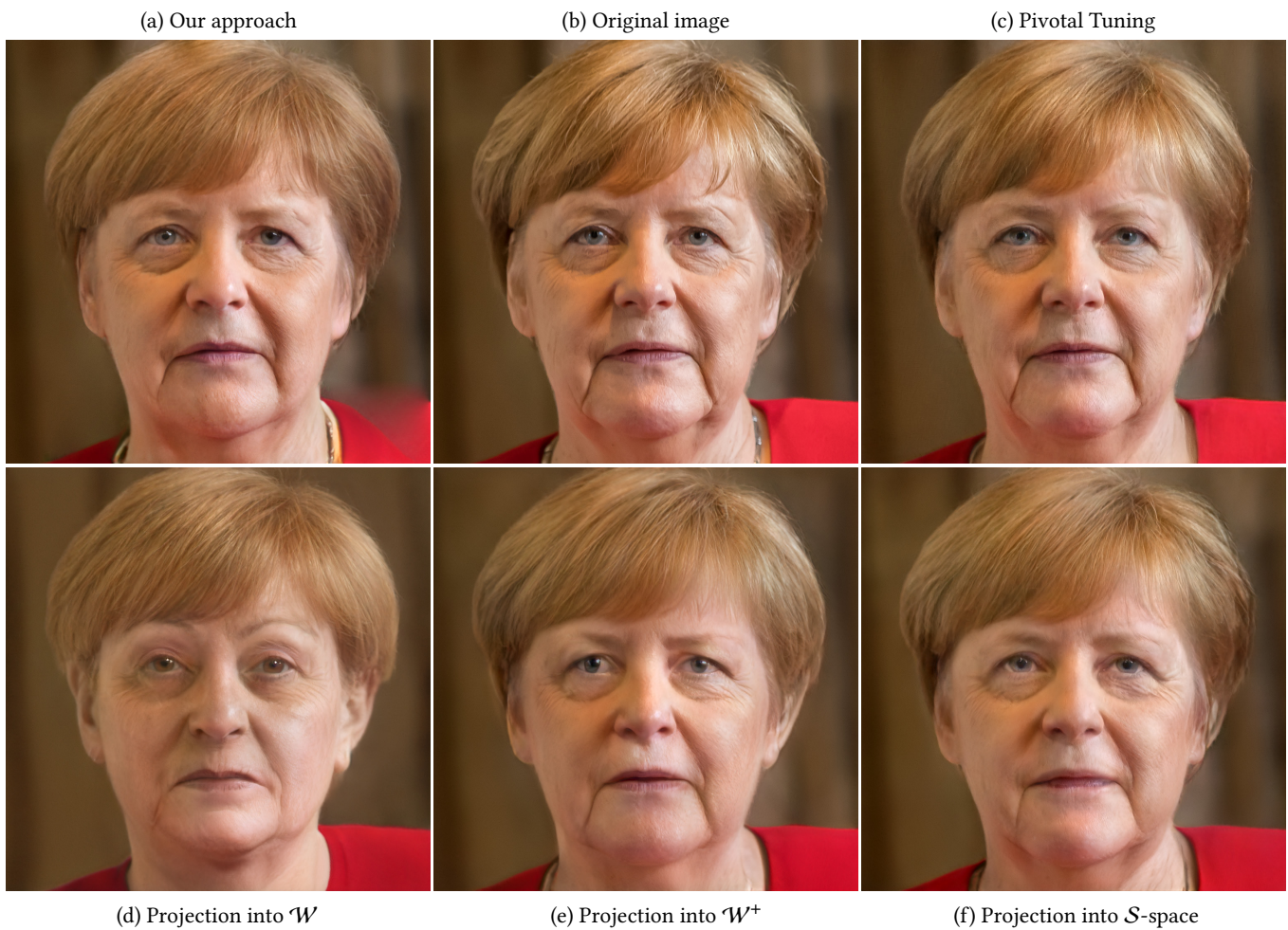}}
\caption{An example of challenging projection task: our approach~(a) used to
reproduce identity of Angela Merkel in the original image~(b). Although Pivotal
Tuning~\cite{Roich21}~(c) achieves better identity preservation it requires
custom StyleGAN v2 model. Other projections based on standard StyleGAN v2 model
such as $\mathcal{W}$~(d), $\mathcal{W}^{+}$~(e),
or~$\mathcal{S}$-space~\cite{Wu21}~(f) tend to alter the identity of Angela
Merkel considerably.} \label{fig:project}
\end{figure*}

\subsection{Segmentation Refinement}\label{sec:lsm}

As an improvement to the projection quality we can refine the segmentation~$S$
during the iterative optimization of~$\mathcal{W}$ or during the editing phase
to avoid segment boundaries to cross salient features~(see~\fg{lsm}). To
perform such a refinement we represent the boundaries of individual segments by
a Level Set~\cite{Osher88} and optimize them to avoid collision with areas of
high difference between adjacent segments.

Let us consider an example shown in~\fg{lsm}. Here we have an initial
segmentation~$S$ that separates eyes from head. In~$S$ eyes are denoted by~$+1$
and head by~$-1$. The boundary between eyes and head segment is denoted
as~$S=0$ (red curve in~\fg{lsm}c). In addition, we define a stopping
function~$F=|I-I_{\mathcal{W}}|$ (see inset in~\fg{lsm}b) where~$I$ is the
original image~(\fg{lsm}a) and~$I_{\mathcal{W}}$ is an image~(\fg{lsm}b)
generated by a latent code~$\mathcal{W}$ optimized over pixels where~$S=+1$
(interior of red curve in~\fg{lsm}c). The Level Set method then iteratively
refines~$S$ by using the following update rule:
$$S' = S + \Delta t \cdot F \cdot || \nabla S ||,$$
where~$\Delta t$ is a small step in time (a new boundary can always be
extracted by setting~$S'=0$). The boundary of~$S$ moves faster at pixels~$p$
where the value of stopping function~$F(p)$ is high (yellow pixels in the inset
of~\fg{lsm}b) otherwise stays put or move slowly (blue pixels). Thanks to this
property the resulting optimized boundary~(\fg{lsm}d) avoids pixels which would
otherwise introduce artifacts~(\fg{lsm}c).

\section{Results}

We test our approach on StyleGAN v2~\cite{Karras20} in the domain of faces,
using gradient descent optimization method based on~\cite{Karras19} to obtain
the projected latent codes~$\mathcal{W}$ for each segment. Segmentation maps
used in these results were hand-drawn by users who were interacting with our
system. In~\fg{segments}, we show an example of full-face projection with
eleven segments each corresponding to a specific semantic region (e.g., eyes,
hair, mouth, ears, etc.). Compared to a global projection of~$\mathcal{W}$, we
obtain an image which is much closer in identity to the original image.
Moreover, the segments are nicely stitched without visible seams or
discontinuities. A more challenging projection result is present
in~\fg{project}. In this case project on~$\mathcal{W}$~(\fg{project}d),
$\mathcal{W}^{+}$~(\fg{project}e),
$\mathcal{S}$-space~\cite{Wu21}~(\fg{project}f), and Pivotal
Tuning~\cite{Roich21}~(\fg{project}c) were used to match the identity in the
target photo~(\fg{project}b). Our approach~(\fg{project}a) performs
substantially better than techniques which do not require custom model, i.e.,
$\mathcal{W}$~(\fg{project}d), $\mathcal{W}^{+}$~(\fg{project}e), and
$\mathcal{S}$-space~(\fg{project}f). Although Pivotal Tuning achieves slightly
better identity preservation than our techniques, it requires custom fine-tuned
StyleGAN v2 model which can be difficult to manipulate.

In~\fg{smile} we show a smile enhancement edit. While projecting
into~$\mathcal{W}^{+}$ achieves identity on par with our projection, the result
of changing the smile through a latent code manipulation in $\mathcal{S}$-space
is far from looking realistic. Contrary, $\mathcal{S}$-space manipulation
performed per-segment in our method retains more realism in the final image. An
example of incremental editing scenario can be seen in~\fg{edits}, where a user
selectively draws segments to change desired attributes of a real face. First,
the gaze is changed through changes in $\mathcal{S}$-space, then the smile is
enhanced via a move in the~$\mathcal{W}$ space, and so on.

Interestingly, our method can even be applied to out-of-domain images, such
as paintings in~\fg{artworks} where we perform changes on an artwork using a
network trained exclusively on real photographs of faces. Because the segments
are relatively small, the projection is still feasible and subsequent edits
look convincing.

\begin{figure*}
\centerline{\def\svgwidth{\linewidth}
\begingroup%
  \makeatletter%
  \providecommand\color[2][]{%
    \errmessage{(Inkscape) Color is used for the text in Inkscape, but the package 'color.sty' is not loaded}%
    \renewcommand\color[2][]{}%
  }%
  \providecommand\transparent[1]{%
    \errmessage{(Inkscape) Transparency is used (non-zero) for the text in Inkscape, but the package 'transparent.sty' is not loaded}%
    \renewcommand\transparent[1]{}%
  }%
  \providecommand\rotatebox[2]{#2}%
  \newcommand*\fsize{\dimexpr\f@size pt\relax}%
  \newcommand*\lineheight[1]{\fontsize{\fsize}{#1\fsize}\selectfont}%
  \ifx\svgwidth\undefined%
    \setlength{\unitlength}{2303.99979238bp}%
    \ifx\svgscale\undefined%
      \relax%
    \else%
      \setlength{\unitlength}{\unitlength * \real{\svgscale}}%
    \fi%
  \else%
    \setlength{\unitlength}{\svgwidth}%
  \fi%
  \global\let\svgwidth\undefined%
  \global\let\svgscale\undefined%
  \makeatother%
  \begin{picture}(1,0.35513089)%
    \lineheight{1}%
    \setlength\tabcolsep{0pt}%
    \put(0,0){\includegraphics[width=\unitlength,page=1]{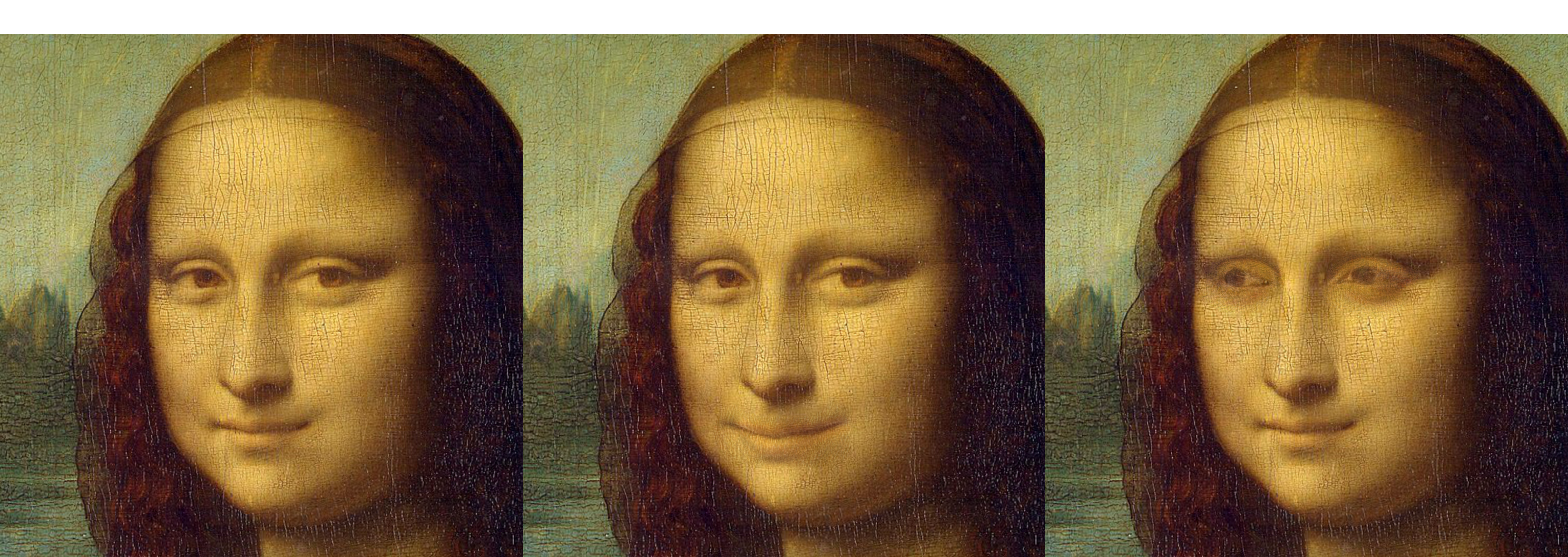}}%
    \put(0.020,0.31){\color[rgb]{1,1,1}\makebox(0,0)[t]{\lineheight{1.25}\smash{\begin{tabular}[t]{c}(a)\end{tabular}}}}%
    \put(0.355,0.31){\color[rgb]{1,1,1}\makebox(0,0)[t]{\lineheight{1.25}\smash{\begin{tabular}[t]{c}(b)\end{tabular}}}}%
    \put(0.689,0.31){\color[rgb]{1,1,1}\makebox(0,0)[t]{\lineheight{1.25}\smash{\begin{tabular}[t]{c}(c)\end{tabular}}}}%
  \end{picture}%
\endgroup%
}
\centerline{\def\svgwidth{\linewidth}
\begingroup%
  \makeatletter%
  \providecommand\color[2][]{%
    \errmessage{(Inkscape) Color is used for the text in Inkscape, but the package 'color.sty' is not loaded}%
    \renewcommand\color[2][]{}%
  }%
  \providecommand\transparent[1]{%
    \errmessage{(Inkscape) Transparency is used (non-zero) for the text in Inkscape, but the package 'transparent.sty' is not loaded}%
    \renewcommand\transparent[1]{}%
  }%
  \providecommand\rotatebox[2]{#2}%
  \newcommand*\fsize{\dimexpr\f@size pt\relax}%
  \newcommand*\lineheight[1]{\fontsize{\fsize}{#1\fsize}\selectfont}%
  \ifx\svgwidth\undefined%
    \setlength{\unitlength}{2303.99979238bp}%
    \ifx\svgscale\undefined%
      \relax%
    \else%
      \setlength{\unitlength}{\unitlength * \real{\svgscale}}%
    \fi%
  \else%
    \setlength{\unitlength}{\svgwidth}%
  \fi%
  \global\let\svgwidth\undefined%
  \global\let\svgscale\undefined%
  \makeatother%
  \begin{picture}(1,0.66666669)%
    \lineheight{1}%
    \setlength\tabcolsep{0pt}%
    \put(0,0){\includegraphics[width=\unitlength,page=1]{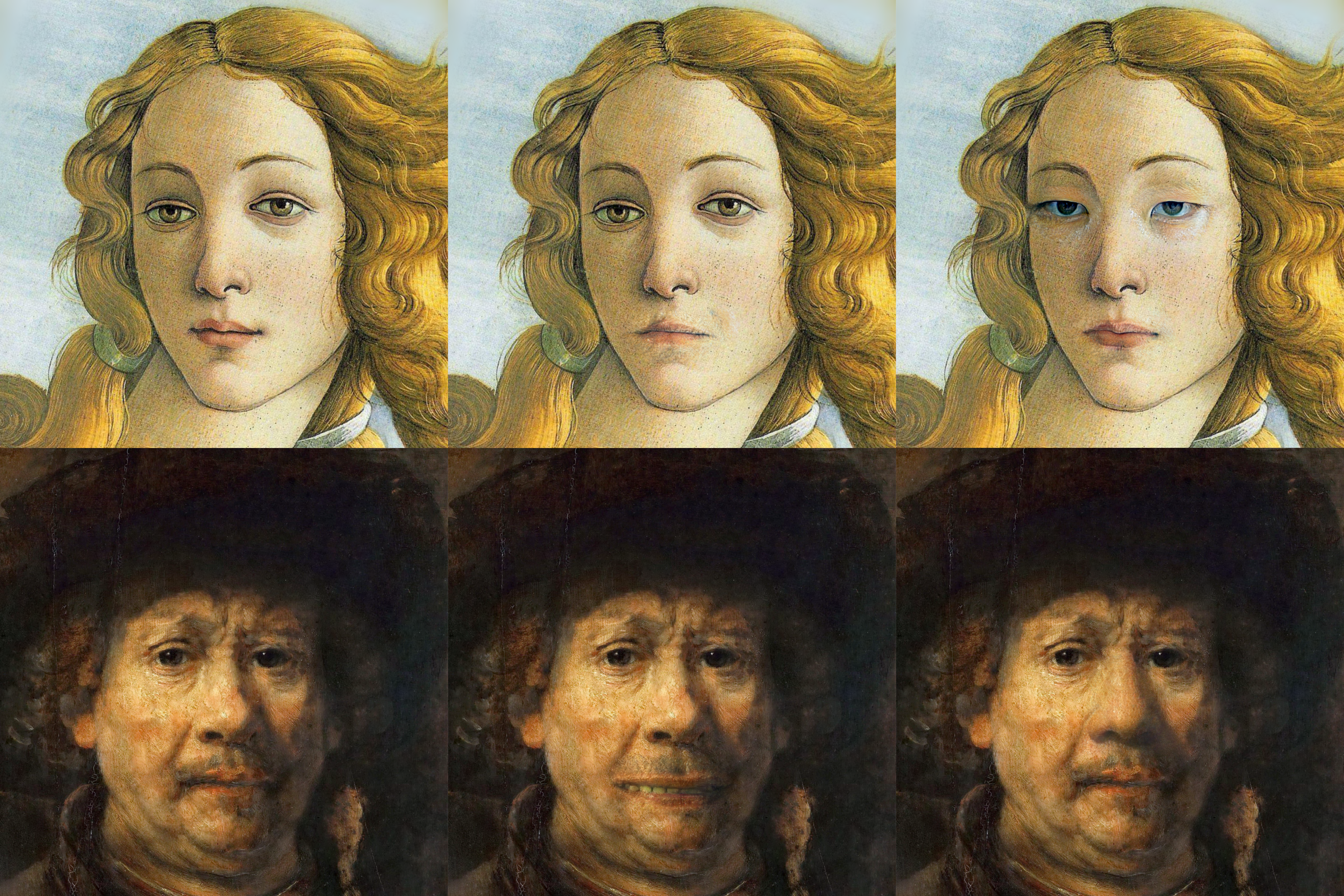}}%
    \put(0.00909629,0.645){\color[rgb]{0,0,0}\makebox(0,0)[lt]{\lineheight{1.25}\smash{\begin{tabular}[t]{l}(d)\end{tabular}}}}%
    \put(0.34073725,0.645){\color[rgb]{0,0,0}\makebox(0,0)[lt]{\lineheight{1.25}\smash{\begin{tabular}[t]{l}(e)\end{tabular}}}}%
    \put(0.67416362,0.645){\color[rgb]{0,0,0}\makebox(0,0)[lt]{\lineheight{1.25}\smash{\begin{tabular}[t]{l}(f)\end{tabular}}}}%
    \put(0.00909629,0.310){\color[rgb]{1,1,1}\makebox(0,0)[lt]{\lineheight{1.25}\smash{\begin{tabular}[t]{l}(g)\end{tabular}}}}%
    \put(0.34073725,0.310){\color[rgb]{1,1,1}\makebox(0,0)[lt]{\lineheight{1.25}\smash{\begin{tabular}[t]{l}(h)\end{tabular}}}}%
    \put(0.67416362,0.310){\color[rgb]{1,1,1}\makebox(0,0)[lt]{\lineheight{1.25}\smash{\begin{tabular}[t]{l}(i)\end{tabular}}}}%
  \end{picture}%
\endgroup%
}
\caption{Thanks to the lower number of constraints, our technique can achieve
edits that would normally require StyleGAN trained on a different domain. Here,
we perform edits on famous artworks using StyleGAN v2 trained solely on real
faces: (a)~Da Vinci's Mona Lisa, (b)~more pronounced smile, (c)~change in the
gaze direction, (d)~Botticelli's The Birth of Venus, (e)~change in the mouth
expression, (f)~different shape of eyes, (g)~Rembrandt's Little Self-portrait,
(h)~changing mouth expression, (i)~different shape of the nose.}
\label{fig:artworks}
\end{figure*}
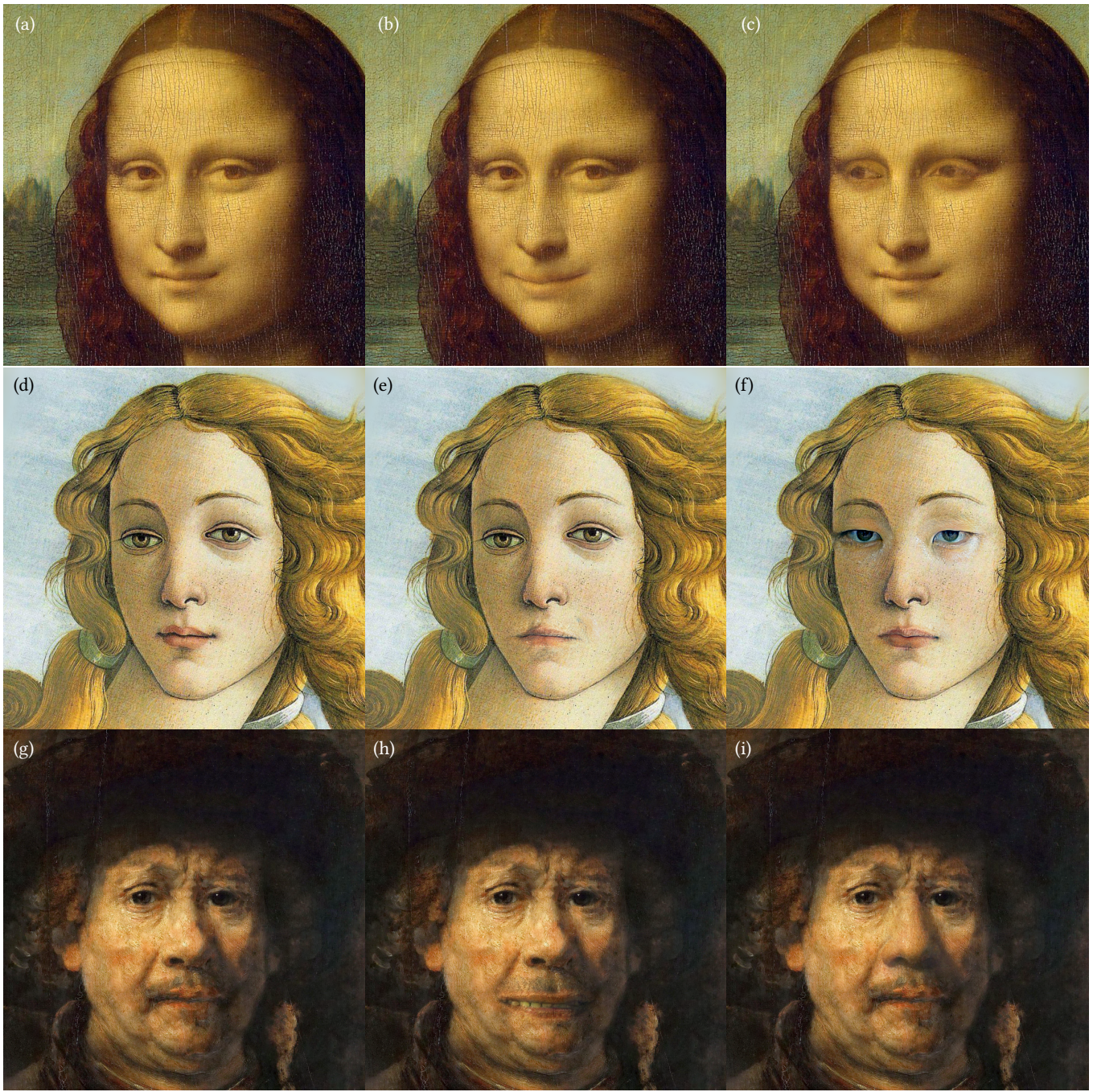

\begin{figure*}[ht]
\centerline{\def\svgwidth{\linewidth}\input{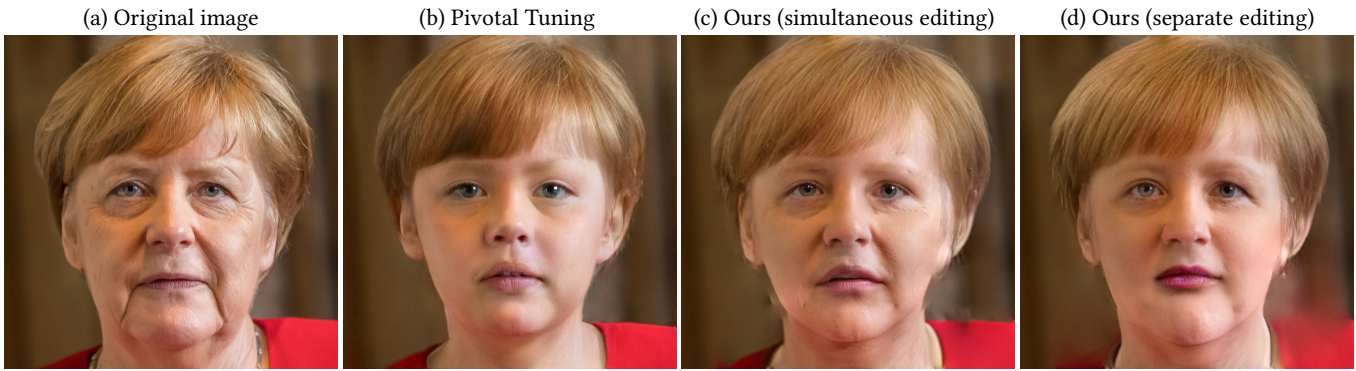}}
\caption{Limitation---when a more dramatic edit in the latent space is applied
on the original image~(a), state-of-the-art global methods such as Pivotal
Tuning~\cite{Roich21}~(b) would produce consistent output while our approach
may introduce inconsistencies between individual segments when edited
simultaneously~(c). This limitation can be alleviated by editing each segment
separately~(d). Note, how our approach better preserves identity after editing.}
\label{fig:age_edit}
\end{figure*}

\section{Limitations}

Despite the fact the current trend in the literature is to perform edits that
have global impact on the entire image, we argue that in practice, users also
often do selective, smaller edits for which our method is highly beneficial.
Although, our approach supports global editing as well (by changing the latent
vectors in all segments simultaneously), it could encounter difficulties when
the change is larger as inconsistencies between segments may became apparanet
(see~\fg{age_edit}c). Smaller mismatch can partially be resolved by the
automatic boundary refinement discussed in~\sec{lsm}, for more dramatic changes
separate editing of latent code in each segment would yield better
results~(\fg{age_edit}d).

\section{Conclusion}

We presented an approach to image editing based on generative adversarial
networks that subdivide the input image into a set of segments for which the
corresponding latent vectors are retrieved separately. In contrast to global
methods that encode the entire image, our local inversion enables more accurate
reconstructions, leading, e.g., to better identity preservation in facial
images. We demonstrated the utility of our technique in numerous practical
scenarios where previous methods may encountered difficulties. We believe that
the complementary nature of our approach finds its application potential in
modern image editing tools.

\bibliographystyle{ACM-Reference-Format}
\bibliography{main}


\end{document}